\def\year{2022}\relax
\documentclass[letterpaper]{article} 
\usepackage{aaai22}  
\usepackage{times}  
\usepackage{helvet}  
\usepackage{booktabs}       

\usepackage{courier}  
\usepackage[hyphens]{url}  
\usepackage{multicol}
\usepackage{graphicx} 
\usepackage{natbib}  
\usepackage{caption} 

\DeclareCaptionStyle{ruled}{labelfont=normalfont,labelsep=colon,strut=off} 
\usepackage{algorithm}
\usepackage{algorithmic}
\usepackage{hyperref}
\usepackage{newfloat}
\usepackage{listings}
\floatstyle{ruled}
\newfloat{listing}{tb}{lst}{}
\floatname{listing}{Listing}
\usepackage{amssymb}

\title{PrivateMail: Supervised Manifold Learning of Deep Features\\ With Privacy for Image Retrieval
} 
\author {
    Praneeth Vepakomma,
    Julia Balla, 
    Ramesh Raskar
}
\affiliations {
    Massachusetts Institute of Technology\\
    vepakom@mit.edu
}
\usepackage{amsthm}
\usepackage{amsfonts}
\usepackage{mathtools}
\usepackage[algo2e]{algorithm2e}
\usepackage{wrapfig}
\newtheorem{definition}{Definition}
\newtheorem{theorem}{Theorem}
\newtheorem{lemma}{Lemma}

\newenvironment{psketch}{
  \proof}{\endproof}

\DeclareMathOperator{\Diag}{Diag}
\DeclareMathAlphabet{\mathcal}{OMS}{cmsy}{m}{n}
\DeclareMathOperator{\Tr}{Tr}

\begin{document}
\maketitle
\vspace{-3mm}
\begin{abstract}
Differential Privacy offers strong guarantees such as immutable privacy under any post-processing. In this work, we propose a differentially private mechanism called PrivateMail for performing supervised manifold learning. We then apply it to the use case of private image retrieval to obtain nearest matches to a client's target image from a server's database. PrivateMail releases the target image as part of a differentially private manifold embedding. We give bounds on the global sensitivity of the manifold learning map in order to obfuscate and release embeddings with differential privacy inducing noise. We show that PrivateMail obtains a substantially better performance in terms of the privacy-utility trade off in comparison to several baselines on various datasets. We share code for applying PrivateMail at \url{http://tiny.cc/PrivateMail}. 
\end{abstract}
\vspace{-2mm}
\section{Introduction}
Privacy preserving computation enables distributed hosts with `siloed' away data to query, analyse or model their sensitive data and share findings in a privacy preserving manner. As a motivating problem, in this paper we focus on the task of privately retrieving nearest matches to a client's target image with respect to a server's database of images.  Consider the setting where a client would like to obtain the k-nearest matches to its target from an external distributed database. State of the art image retrieval machine learning models such as \citep{cvpr20_tutorial_image_retrieval,chen2021deep,zhou2017recent,dubey2020decade} exist for feature extraction pior to obtaining the neighbors to a given match in the learnt space of deep feature representations. Unfortunately, this approach is not private. The goal of our approach is to be able to use these useful features for the purpose of image retrieval in a manner, that is formally differentially private. The seminal idea for a mathematical notion of privacy, called differential privacy, along with its foundations is introduced quite well in \cite{dwork2014algorithmic}. 
In our approach, we geometrically embed the image features via a supervised manifold learning query that we propose. Our query falls within the framework of supervised manifold learning as formalized in \cite{vural2017study}. We then propose a differentially private mechanism to release the outputs of this query. The privatized outputs of this query are used to perform the matching and retrieval of the nearest neighbors in this privatized feature space. 
Differential privacy aims to prevent membership inference attacks \citep{shokri2017membership,truex2018towards,li2020label,song2019membership, shi2020over}. It has been shown that differential privacy mechanisms can also prevent reconstruction attacks under a constraint on the level of utility that can be achieved as shown in \citep{dwork2017exposed,garfinkel2018understanding}. 
Currently cryptographic methods for the problem of information retrieval were studied in works like \citep{xia2015towards}. These methods ensure to protect the client's data via homomorphic encryption and oblivious transfer. However, they also come with an impractical trade-off of computational scalability, especially when the size of the server's database is large and the feature size is high-dimensional as is always the case in practice \citep{elmehdwi2014secure,lei2019seceqp,yao2013secure}. 

\begin{figure*}[!h]
    \centering
    \includegraphics[scale=0.75]{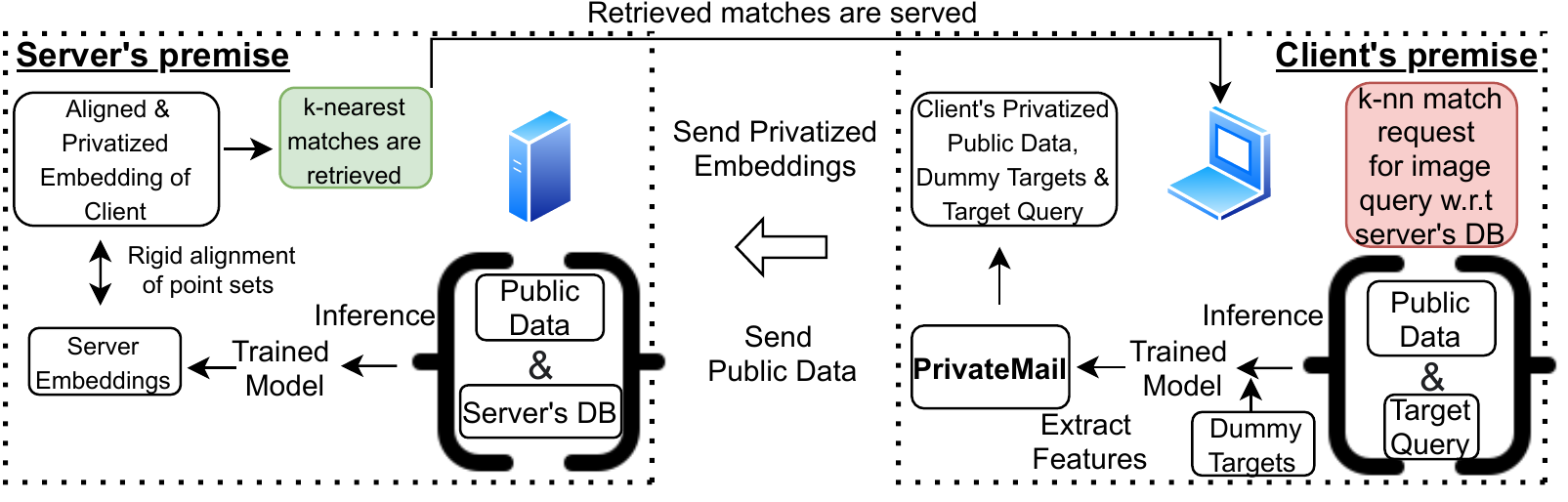}
    \caption{This illustration shows the lifecycle of interactions between client and server side entities for private image retrieval. The interaction starts from the red bubble on the right. At first, the client and server train a good on-premise machine learning model that is tailored for image retrieval. The client extracts features from this model on the target query image, dummy targets that are only known to the client as well as a public dataset known to the client and server. The extracted features go through the proposed Private-Mail for embedding them via locally differentially private supervised manifold learning. These private embeddings are aligned at server prior to performing the nearest-neighbor retrieval of matches that are served back to the client. The privatized representation of public dataset is used as an anchor in order to align the feature embeddings between the client and server.}
    \label{fig:my_label}
\end{figure*}

\subsection{Motivation}
\textbf{1.} Currently available differential privacy solutions for biometric applications where content based matching of records is performed \citep{steil2019privacy,chamikara2020privacy} is based on a small number of hand-crafted features. 
We instead consider state of the art feature extraction used by recent deep learning architectures specialized for image retrieval such as \citep{jun2019combination}. We privatize these features and share them in the form of differentially private embeddings that are in turn used for the image retrieval task.
    \\\textbf{2.} Cryptographic methods with strong security guarantees are currently not scalable computationally, for secure k-nn queries \citep{elmehdwi2014secure,lei2019seceqp,yao2013secure} especially when the server-side database is large as is typically the case in real-life scenarios.    
\vspace{-3mm}
\section{Contributions} 

    \textbf{1.} The main contibution of our paper is a differentially private method called \textit{PrivateMail} for private release of outputs from a supervised manifold learning query that embeds data into a lower dimension. We test our scheme for differentially private `content based image retrieval', where the matches to a target image requested by a client are retrieved from a server's database while maintaining differential privacy. 
    \\\textbf{2.} We show a substantial improvement in the utility-privacy trade-off of our embeddings over 5 existing baselines.
    \\\textbf{3.} The supervised manifold learning query that we propose to geometrically embed features extracted from deep networks is novel in itself. That said, we would only consider this as a secondary contribution to this paper. 

\vspace{-3mm}\section{Related work}\vspace{-1.7mm}
\textbf{Non-private image search and retrieval:}
Current state of the art pipelines for content based image retrieval under the non-private setting are fairly matured and based on nearest neighbor queries performed over specialized deep feature representations of these images. The query image and the database of images are compared in this learnt representation space. A detailed set of tutorials and surveys on this problem in the non-private setting is provided in \citep{cvpr20_tutorial_image_retrieval, chen2021deep,zhou2017recent,dubey2020decade}. 
\\ \textbf{Private manifold learning:} There have been recent developments in learning private geometric embeddings 
with differentially private unsupervised manifold learning. Notable examples include distributed and differentially private version of t-SNE \cite{van2008visualizing} called DP-dSNE \citep{saha2020dsne,saha2021privacy} and \cite{arora2019differentially} for differentially private Laplacian Eigenmaps \cite{belkin2003laplacian,belkin2007convergence}. Furthermore, the work in \citep{choromanska2016differentially} provides a method for differentially private random projection trees to perform unsupervised private manifold learning. However, none of these works consider differentially private manifold learning in the supervised setting that we explore in this paper. We show a substantial improvement in privacy-utility trade-offs of the supervised manifold embedding approach over existing baselines that include private and non-private methods in the supervised and unsupervised paradigms.
 \vspace{-3mm}\section{Approach}
Motivated by the supervised manifold learning framework in \citep{vural2017study} that is based on a difference of two unsupervised manifold learning objectives, we present an iterative update to efficiently optimize it. We refer to this iterative optimization as the \textit{supervised manifold learning query (SMLQ)}. We then provide a privacy mechanism called \textit{PrivateMail} to perform this supervised manifold learning query with a guarantee of differential privacy. To do that, we derive the sensitivity of our query that is required to calibrate the amount of noise needed to attain differential privacy. As part of experimental results, we apply our approach to a novel task of differentially private image retrieval, that has not been well-studied in current literature as opposed to the non-private image retrieval task which is a widely studied problem. 
\begin{table}[!htbp]
\centering
\resizebox{5cm}{!}{
\begin{tabular}{|l|l|}
\hline
Notation                 & Description                    \\ \hline
$n$                      & Sample size                    \\ \hline
$d$                      & Data dimension                 \\ \hline
$k$                      & Embedded dimension             \\ \hline
$\mathbf{X}_{n\times d}$ & Data matrix                    \\ \hline
$\mathbf{Y}_{n\times 1}$ & Labels                         \\ \hline
$f$                      & Manifold learning map          \\ \hline
$\sigma$                 & Gaussian kernel bandwidth      \\ \hline
$\sigma_q$               & std. dev. of entries in $\mathbf{Q}$                               \\ \hline
$\alpha$                 & regularization in $\mathbf{L_X-\alpha L_Y}$                               \\ \hline
$\mathbf{Q}$             & $Q_{i,j} \sim N(0,\sigma_q^2)$ \\ \hline
\end{tabular}
}
\caption{Notations}
\end{table}
\vspace{-6mm}

\section{Moving from unsupervised to supervised manifold learning}

We first briefly introduce some preliminaries for unsupervised manifold learning in order to build upon it to introduce supervised manifold learning.
\subsection{Preliminaries for unsupervised manifold learning} This problem is a discrete analogue of the continuous problem of learning a map $f:\mathcal{M}\mapsto \mathbb{R}^k$ from a smooth, compact high dimensional Riemannian manifold such that for any two points $x_1,x_2$ on $\mathcal{M}$, the geodesic distance on the manifold $d_{\mathcal{M}}(x_1,x_2)$ is approximated by the Euclidean distance $\Vert f(x_1) - f(x_2)\Vert$ in $\mathbb{R}^k$. Different manifold learning techniques vary in their tightness of this approximation on varying datasets. Manifold learning techniques like Laplacian Eigenmaps \cite{belkin2005towards}, Diffusion Maps \cite{coifman2006diffusion} and Hessian Eigenmaps \cite{donoho2003hessian} aim to find a tighter approximation by trying to minimize a relevant bounding quantity \textbf{B} such that $\Vert f(x_1) -f(x_2)\Vert \leq \textbf{B}\cdot d_{\mathcal{M}}(x_1,x_2) + o(d_{\mathcal{M}}(x_1,x_2))$. 
Different techniques propose different possiblilities for such a $\textbf{B}$. For example, Laplacian Eigenmaps uses $\textbf{B}=||\nabla{f}(x_1)||$ for which it is shown that this relation holds as
\begin{equation*}
    \resizebox{0.95\hsize}{!}{$ 
     {||f(x_1)-f(x_2)||\leq ||\nabla{f}(x_1)||\cdot||x_1-x_2||+ o(||x_1-x_2||)}
$} 
\end{equation*}
Hence, controlling $||\nabla{f}||_{L^2(M)}$ preserves geodesic relations on the manifold in the Euclidean space after the embedding. \subsection{From continuous to discrete }This quantity of  $||\nabla{f}||_{L^2(M)}$ in the continuous domain can be optimized via chosing the eigenfunctions of the Laplace-Beltrami operator in order to get the optimal embedding. This is explained in a series of papers by \citep{gine2006empirical,belkin2007convergence,jones2008manifold}. From a computational standpoint we note that, for a specific graph defined on all pairs of data points with an adjacency matrix $\mathbf{W_X}$ and corresponding graph Laplacian $\mathbf{L_X}$, the following quantity
\begin{equation}\label{unsupManifEqn}
    \Sigma_{i,j}(||\mathbf{f(X_{i})} -\mathbf{f(X_{j})}||^2 \cdot [\mathbf{W_X}]_{ij})= \Tr(\mathbf{f(X)}^{T}\mathbf{L_X}\mathbf{f(X)})
\end{equation} is the discrete version of $||\nabla{f}||^{2}_{L^2(\mathcal{M})}$ under the assumption that the dataset $\mathbf{X}$ is a sample lying on the manifold $\mathcal{M}$. Here, $\mathbf{f(X_{i}})$ and $\mathbf{f(X_{j}})$ refer to the $k$ dimensional real-valued output of the manifold learning map $f$ at two single points represented by $i$ and $j$ rows in the data matrix $\mathbf{X_{n \times d}}$. Similarly, $\mathbf{f(X)}$ refers to mapping the points indexed by each row in $\mathbf{X}$ to $\mathbb{R}^k$. That is, the output of $\mathbf{f(X)}$ is a real-valued matrix of dimension $n\times k$. Therefore, the equivalent solution to map $\{X_{1},...X_{n}\}\subset \mathbb{R}^d$ while preserving local neighborhood into $\{f(X_{1}),...f({X_{n}})\}\subset \mathbb{R}^k$ is to minimize this objective function in \eqref{unsupManifEqn} for a specific graph Laplacian $\mathbf{L_X}$ that we describe below. This popular graph Laplacian, under which the above results were studied is that of graphs whose adjacency matrices are represented by the Gaussian kernel given by 
 \begin{equation}\label{LEqn1}
\mathbf{L(\mathbf{X}},\sigma)_{ik}=\left\{
\begin{matrix} 
\sum_{k \ne i} e^{( - \frac{\Vert \mathbf{X_i}- \mathbf{X_k}\Vert^2}{ \sigma} )} & \mbox{if}\ i = k \\
-e^{(-\frac{ \Vert \mathbf{X_i} - \mathbf{X_k}\Vert^2}{ \sigma})} & \mbox{if}\ i \neq k
\end{matrix}\right\} \end{equation} 
where the scalar $\sigma$ in here is also referred to as kernel bandwidth. The seminal work in
\citep{gine2006empirical, belkin2005towards, belkin2007convergence} showed that this discrete Graph Laplacian converges to the Laplace-Beltrami operator.  
Minimizing this objective of Equation \ref{unsupManifEqn} under the constraint $ \Tr(\mathbf{f(X)^{T}Df(X))=I}$ where $\mathbf{I}$ is identity matrix, to avoid a trivial solution of  ${\mathbf{\Tr(f(X)^{T}L_Xf(X))}=0}$ is equivalent to setting the solution for the embedding $\mathbf{f(X)}$ to be the $d$ smallest eigenvectors of $\mathbf{L_X}$.
\subsection{Supervised manifold learning queries (SMLQ)} 
\begin{figure}
    \centering{\includegraphics[width=0.48\textwidth]{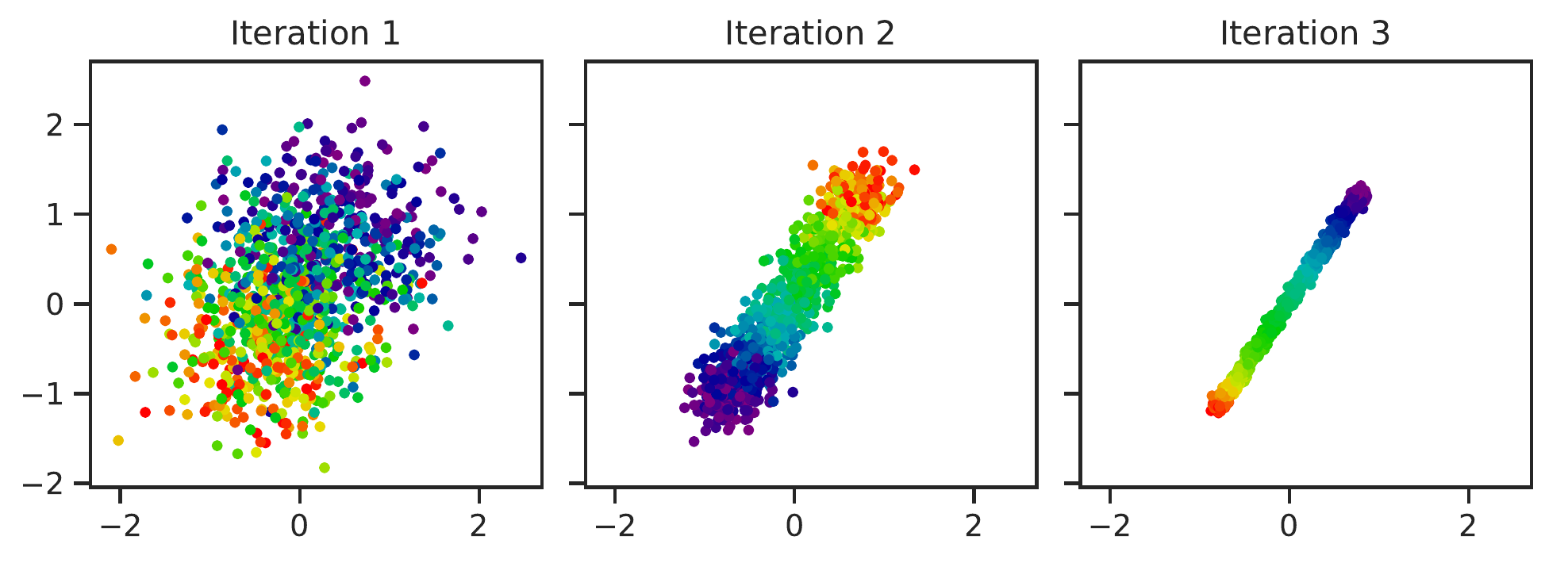}}
    \caption{Embeddings of our supervised manifold learning query on CUB-200-2011 for 3 iterations with input features extracted from state-of-the-art CGD \citep{jun2019combination} deep image retrieval architecture with ResNet 50 backbone and G type global descriptors. The colors indicate different class labels. We show that these embeddings preserve information about the class separation and the locality structure required for classification.}
    \label{fig:2}
\end{figure}

It has been shown in \citep{vural2017study} that this formulation for unsupervised manifold learning of minimizing equation \eqref{LEqn1} can be extended to the case of supervised manifold learning by posing the objective function as a difference of the terms in \eqref{unsupManifEqn} as shown below.\begin{equation} \label{smfeqn}
v(\mathbf{f(X)}) = Tr(\mathbf{f(X)^T L_X f(X)}) - \alpha Tr(\mathbf{f(X)^T L_Yf(X)})\end{equation} Note that the formula for computing $\mathbf{L_Y}$ over $\mathbf{Y}$, is the same as the one used in \eqref{LEqn1} to compute $\mathbf{L_X}$ from $\mathbf{X}$. They provide results explaining the effect of optimizing such a loss for the purposes of learning an embedding $\mathbf{f(X)}$ for supervised learning. Their results are agnostic to the choice of neighborhood graphs defined on $\mathbf{X,Y}$ to obtain the corresponding Laplacians used in this objective. An example for such an embedding when applied to features extracted from state-of-the-art CGD (Jun et al. 2019) deep image retrieval architecture with ResNet 50 backbone is shown in Figure \ref{fig:2}.
\subsubsection{Separation-regularity trade-off}
The intuition is that since equation \eqref{smfeqn} is a discrete version of a difference of terms of the kind in \eqref{unsupManifEqn}, therefore this formulation looks for a function that has a slow variation on the manifold $\mathcal{M}_X$ in order to smoothly preserve neighborhood relations between the input features. It does this while ensuring the function has a fast variation on a manifold $\mathcal{M}_Y$ with regards to $\mathbf{Y}$, therefore encouraging larger separation with regards to the label manifold. Therefore, this second term acts as a regularizer to make sure similar features are not embedded way closer than needed. This is mathematically substantiated by Theorem 9 in \cite{vural2017study} (restated in the Appendix D) as it shows that this regularization is required in order to minimize the generalization error of a classifier applied on the output of supervised manifold learning obtained via minimization of equation \eqref{smfeqn} for any choice of positive semidefinite $\mathbf{L_X,L_Y}$.

\begin{theorem} 
For a fixed $\alpha$, the iterate \begin{equation}
     \label{superManifQuery}
\mathbf{X_t = \frac{Diag(L_X)^{-1}}{2}[\alpha L_Y - L_X]X_{t-1} + X_{t-1}}\end{equation} monotonically minimizes the objective $$v(\mathbf{X}_t) = Tr(\mathbf{\mathbf{X}_t^T L_X \mathbf{X}_t}) - \alpha Tr(\mathbf{\mathbf{X}_t^T L_Y \mathbf{X}_t})$$
\end{theorem}
\begin{psketch}
The full proof along with the required background is in, appendix \ref{AppdxA}. The proof strategy involves using the majorization-minimization \citep{hunter2004tutorial,lange2016mm,zhou2019mm} procedure in order to obtain this iterative update. We first derive a majorization function, which always upper bounds the objective everywhere except at the current iterate, where it touches it. We then note that this majorization function is a sum of convex and concave functions. This makes the minimization of the majorization function to be equivalent to using the concave-convex procedure \cite{yuille2002concave}. As the update is based on majorization-minimization (MM) and CCCP which itself is a special case of MM, it thereby guarantees monotonic convergence \cite{hunter2004tutorial}. We refer to this iterate as the \textit{Supervised Manifold Learning Query (SMLQ)} and the rest of the paper focuses on releasing the outputs of SMLQ with differential privacy.
\end{psketch} 
\begin{figure}
    \centering
  \includegraphics[width=0.40\textwidth]{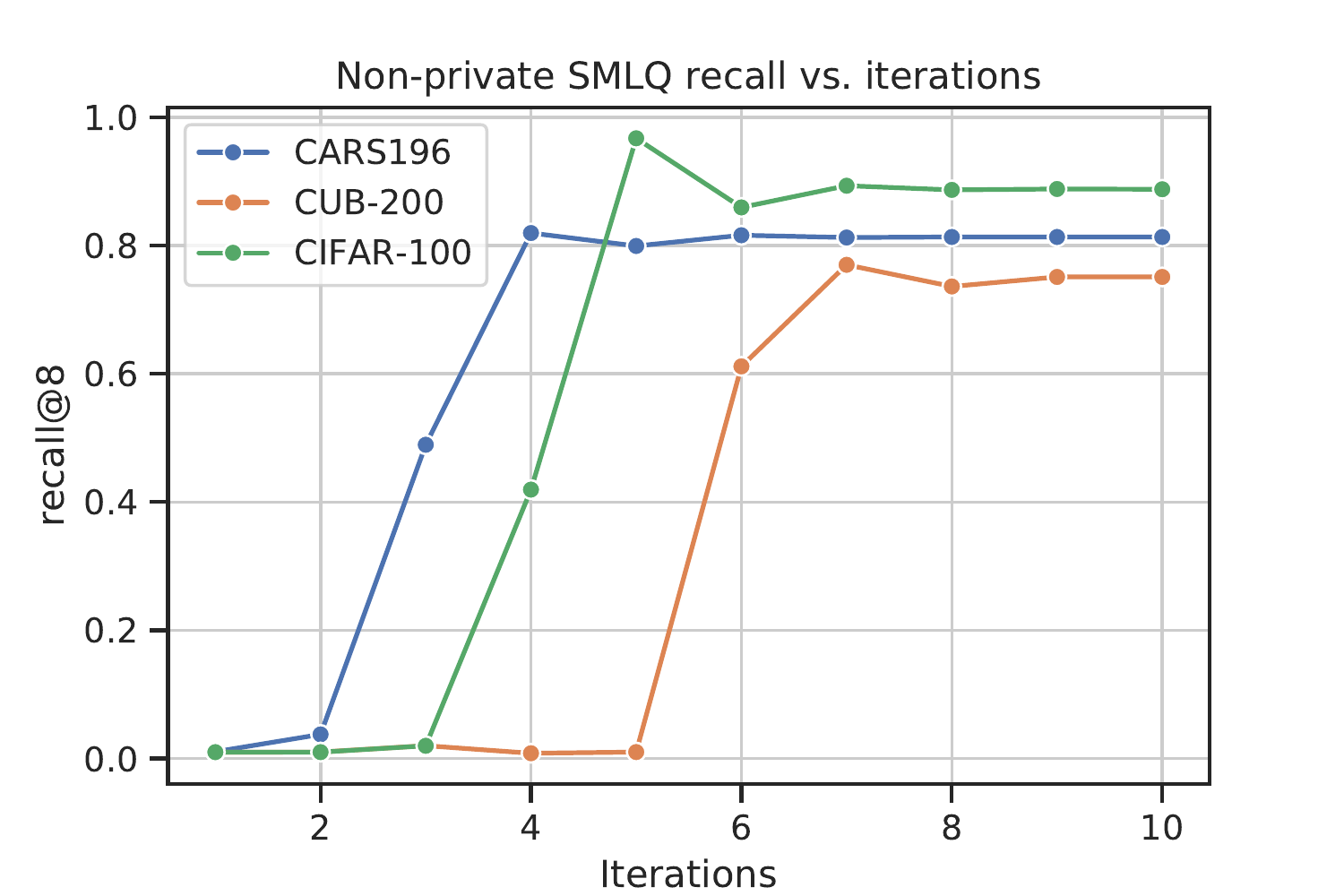}
  \caption{The convergence of our SMLQ across three datasets is shown with respect to image recall based on feature embeddings over the iterations. All three datasets reasonably converge in as quick as $7$ iterations. The image recall metric is discussed in the Experiments section.}
    \label{fig:3}
  \end{figure}
As shown in Figure \ref{fig:3}, our iterative update converges in just $5$ to $7$ iterations to embed deep feature representations needed for an image retrieval task tested on $3$ datasets as further detailed in the experimental section.
\subsubsection{Complexity analysis}
The graph Laplacian based on the Gaussian kernel in our method is sparse and computing the sparse matrix-vector product for this specific graph Laplacian has been studied to take $\mathcal{O}(n)$ time \cite{nfft}. Since in the term $\mathbf{L_Y X}_{t-1}$, the number of columns in $\mathbf{X}_{t-1}$ is $k$, we have an overall time complexity of  $\mathcal{O}(nk)$ as the addition of $n \times k$ matrices also takes $\mathcal{O}(nk)$. That said, this does not include the complexity required to construct the Laplacian. This has been studied in \cite{sanjeev2001learning}. 


\section{Privatization of the Supervised Manifold Learning Query}
\subsection{Preliminaries} We first share some required preliminaries on differential privacy (DP). Differential privacy guarantees that the presence of a particular record in a dataset does not significantly affect the output of a query on the dataset.

\begin{definition}[($\epsilon, \delta$)-Differential Privacy \shortcite{dwork2014algorithmic}] A randomized algorithm $\mathcal{A}:\mathcal{X} \to \mathcal{Y}$ is ($\epsilon, \delta$)-differentially private if, for all neighboring datasets $\mathbf{X, X'}\in \mathcal{X}$ and for all $S \in \mathcal{Y}$,
$$
\Pr[\mathcal{A}(\mathbf{X})\in S] \leq e^\epsilon \Pr[\mathcal{A}(\mathbf{X'})\in S] + \delta
$$
\end{definition} 
\subsubsection{Post-Processing Invariance}Differential privacy is immune to post-processing, meaning that an adversary without any additional knowledge about the dataset $\mathbf{X}$ cannot compute a function on the output $\mathcal{A}(\mathbf{X})$ to violate the
stated privacy guarantees.
\subsubsection{Gaussian noise mechanism} A query on a dataset can be privatized by adding controlled noise from a predetermined distribution. One popular private mechanism is the Gaussian mechanism \citep{dwork2006gaussmech}, which adds Gaussian noise depending on the query's \textit{sensitivity}.
\begin{definition}[$l_2$-sensitivity] Let $f: \mathcal{X} \to \mathbb{R}^k$. The $l_2$-sensitivity of $f$ is
$$
\Delta_2^{(f)} = \max_{\mathbf{X, X'} \in \mathcal{X}}\|f(\mathbf{X})-f(\mathbf{X'})\|_2
$$
where $\mathbf{X, X'}$ are neighboring databases.
\end{definition} 
\begin{definition}[Gaussian Mechanism \shortcite{dwork2014algorithmic}] Let $f: \mathcal{X} \to \mathbb{R}^k$. The Gaussian mechanism is defined as $
\mathcal{M}_G(\mathbf{X}) = f(\mathbf{X}) + \mathbf{Y}$, where $\mathbf{Y}\sim\mathcal{N}^k(0, \sigma^2)$ with $\sigma \geq \frac{\sqrt{2\ln(1.25\ \delta)}\Delta_2^{(f)}}{\epsilon}$. The Gaussian mechanism is $(\epsilon, \delta)$-differentially private.
\end{definition} We use the Gaussian mechanism to privatize the SMLQ, for which we derive the sensitivity.

\subsection{Derivation of SMLQ sensitivity}

We derive a bound on the  sensitivity for the first iteration of the SMLQ,
$\label{eq:iterate}
    f(\mathbf{X}) = \frac{1}{2} {\Diag(\mathbf{L_X})^{\dagger}}\left[\alpha \mathbf{L_Y} - \mathbf{L_X} \right] \mathbf{Q} + \mathbf{Q}
$, where we initialize $\mathbf{X_{0}}$ to a matrix $\mathbf{Q}$ such that each entry is distributed as $\mathbf{Q}_{ij} \sim \mathcal{N}(0, \sigma_q^2)$, for which $\sigma_q$ is a hyperparameter chosen by the user. It is typical to use random initialization for iterative optimization. 
We also assume that $\mathbf{X} \in \mathbb{R}^{n \times k}$ is normalized to have unit norm rows. Under all possible cases of adding one additional unit norm record to $\mathbf{X}$ to produce a neighboring dataset $\mathbf{\Tilde{X}} \in \mathbb{R}^{(n+1) \times k}$ (denoted by the constraint $d(\mathbf{X},\mathbf{\Tilde{X}})=1$),
the sensitivity of our query is defined as
$\label{eq:GS_def}
    \Delta_2^{(f)} = \max_{\mathbf{X}, \mathbf{\Tilde{X}}: d(\mathbf{X},\mathbf{\Tilde{X}})=1}\|f(\mathbf{X})-f(\mathbf{\Tilde{X}})\|_F
$.
Note that we append an extra row of zeroes to $\mathbf{X}$ and $\mathbf{Y}$ such that the matrix dimensions agree with $\mathbf{\Tilde{X}}$ and $\mathbf{\Tilde{Y}}$ when evaluating $f(\mathbf{X})-f(\mathbf{\Tilde{X}})$. To simplify further calculations, we let $\mathbf{M}$ denote the matrix defined by
\small
\begin{equation}\label{eq:Z}
    \mathbf{M}(\mathbf{X}, \mathbf{\Tilde{X}}) =\Diag(\mathbf{L_X})^{\dagger}\left[\alpha \mathbf{L_Y} - \mathbf{L_X} \right] - \Diag(\mathbf{L_{\Tilde{X}}})^{\dagger}\left[\alpha \mathbf{L_{\Tilde{Y}}} - \mathbf{L_{\Tilde{X}}}\right]
\end{equation}
\normalsize
and let $\mathbf{M_i}$ denote the $i$th row of $\mathbf{M}$. 

\begin{figure*}
\centering
\fbox{
\begin{minipage}{6.5in}
\begin{center}
	$\mathsf{Private}{\mathsf{Mail}}$ 
\end{center}
\begin{enumerate}
	\item {\bf Client's input:} Raw data (or activations) $\mathbf{X}$ normalized to have unit norm rows and integer labels $\mathbf{Y}$, Gaussian kernel bandwidth $\sigma$, regularizing parameter $\alpha$, variance $\sigma_q^2$ for random embedding initialization.
	\item {\bf Client computes embedding:} $
	    \mathbf{X_t = \frac{1}{2} \Diag(L_X)^{\dagger}[\alpha L_Y - L_X]X_{t-1} + X_{t-1}}
$
with initialization $\mathbf{X_0}=\mathbf{Q}$ such that $\mathbf{Q}_{ij} \sim \mathcal{N}(0, \sigma_q^2)$, $\mathbf{L_X}$ and $\mathbf{L_Y}$ are graph Laplacians formed over adjacency matrices upon applying Gaussian kernels to $\mathbf{X,Y}$ with bandwidth $\sigma$ . 
	\item {\bf Client side privatization:} The client takes the following actions:
	\begin{enumerate}
     	\item \textbf{Initialization:} Compute constant $M$ that depends on chosen $\alpha, \sigma$ and data size $n$ as defined in appendix \ref{appendix:GS}.

	\item \textbf{Computation of global-sensitivity:} Compute upper bound on global sensitivity as $\Delta = \frac{M\sqrt{n+1}}{2}\|\mathbf{Q}\|_F$ 
	\item \textbf{Add differentially private noise} Release $\mathbf{X_t}$ with the global sensitivity upper bound in step $3(b)$ via the $(\epsilon,\delta)$- differentially private multi-dimensional Gaussian mechanism: $\mathbf{X_t} + \mathcal{N}^{n \times k}\left(\mu=0, \sigma^2 = \frac{2\ln(1.25/\delta)\cdot \Delta^2}{\epsilon^2}\right)$

	\end{enumerate}

	\end{enumerate}
\end{minipage}
}
\caption{Protocol for the proposed PrivateMail mechanism}
\label{fig:protband}

\end{figure*}
\begin{theorem} \textbf{SMLQ sensitivity  bound}\label{thm:GS} We have that, $ \label{eq:GS_bound}
    \Delta_2^{(f)} \leq \frac{M\sqrt{n+1}}{2}\|\mathbf{Q}\|_F
    $.
    where $M$ is a constant defined in appendix \ref{appendix:GS} such that $M \geq \|\mathbf{M_i}\|$ for all $\mathbf{X}$ and $\mathbf{\Tilde{X}}$.
\end{theorem}

\begin{proof}
    Note that $f(\mathbf{X})-f(\mathbf{\Tilde{X}})$ may be expressed as the product $\frac{1}{2}\mathbf{MQ}$. Thus, by sub-multiplicativity of the Frobenius norm, the global sensitivity is bounded by
    \begin{align}\label{eq:GS_inter}
        \Delta_2^{(f)} &= \max_{\mathbf{X}, \mathbf{\Tilde{X}}: d(\mathbf{X},\mathbf{\Tilde{X}})=1}\left\|\frac{1}{2}\mathbf{MQ}\right\|_F\nonumber\\
        &\leq \frac{1}{2}\|\mathbf{Q}\|_F \cdot \max_{\mathbf{X}, \mathbf{\Tilde{X}}: d(\mathbf{X},\mathbf{\Tilde{X}})=1} \|\mathbf{M}\|_F
    \end{align}
    Since $\|\mathbf{M}\|_F = \sqrt{\sum_{i=1}^{n+1}\|\mathbf{M_i}\|^2}$, then if $M$ is a constant as defined in the theorem, we have $\|\mathbf{M}\|_F \leq \sqrt{\sum_{i=1}^{n+1}{M}^2} = M \sqrt{n+1}$. Substituting this expression into the above inequality, we obtain the bound in the theorem. The derivation of a constant $M$ relies on expanding the definition of the Laplacian matrices in \eqref{superManifQuery} and applying law of cosines for the difference of vectors.
    For the full derivation, see appendix \ref{appendix:GS}.
\end{proof}
The above bound on $\Delta_2^{(f)}$ is computed for the sensitivity parameter when adding differentially private noise to the data embedding. Figure 4 summarizes the procedure for privatization, which we call \textit{PrivateMail}.

\subsection{Private iteration-distribute-recursion framework}
We show that the proposed SMLQ, fortunately can be applied under a specific framework that we propose so that it can be used in conjunction with the post-processing property of differential privacy to its advantage in obtaining a much better trade-off of utility and privacy. In addition, it allows for distributing the work required for completing the iterative embedding across multiple distributed entities while still preserving the privacy. This helps further reduce the computational requirements of the client device, prior to distributing the work. The framework still holds in improving the utility-privacy trade-off even if used without distributing the computation. We notice that the only term that requires accessing the sensitive raw dataset is $\mathbf{L_X}$, but the good thing is that this term does not change over iterations, and hence is not sub-scripted by iteration $t$ as we show in equation \ref{superManifQuery}. Therefore, we first apply our proposed differentially private release of PrivateMail, to just the first iteration. The privately obtained embedding is instead used this time to re-build the graph Laplacian $\mathbf{L_X}$. From the next iteration onwards this modified Laplacian is used instead and the post-processing property of differential privacy now holds as no iteration from now onwards needs access to the raw dataset. For this reason these iterations can as well be continued over the server or another device as opposed to the original client device that runs the first PrivateMail iteration.


\begin{figure*}
        \centering \includegraphics[scale=0.4]{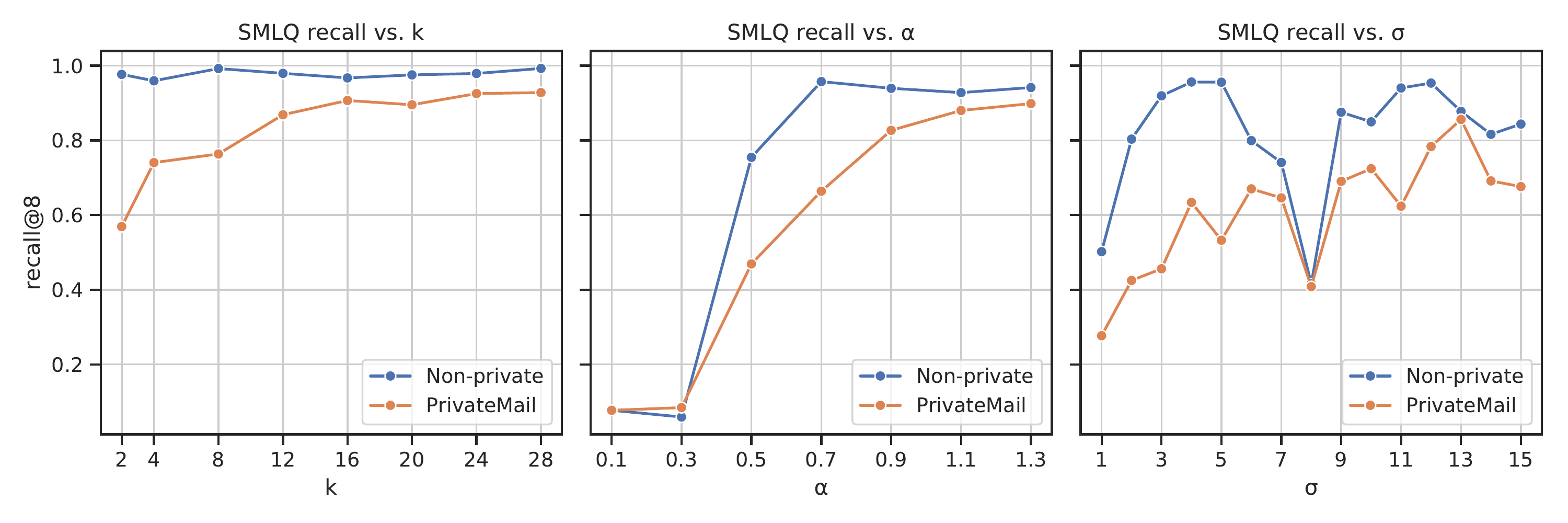}
        \caption{The effect of choice of $k$, $\alpha$, and $\sigma$ on the retrieval performance is shown with respect to the non-private version of SMLQ and the private version: PrivateMail.}
        \label{fig:5}
\end{figure*}

\subsection{PrivateMail for Image Retrieval} We apply the proposed PrivateMail mechanism to the task of private content-based image retrieval, where a client seeks to retrieve the $k$-nearest neighbors of their target image $\mathbf{r}$ from a server's database $\mathcal{S}$ based on the feature embedding of their target which is sent to the server. The objective is to preserve the privacy of the client's target image. We assume the setting in which the client and server have access to a relevant public database $\mathcal{P}$ of images. We propose a differentially private image retrieval algorithm where we first generate feature vectors for $\mathbf{r}$, $\mathcal{P}$, and $\mathcal{S}$ using any feature extraction model of choice. We then generate low-dimensional embeddings for these features using the SMLQ in \eqref{superManifQuery}. Since the query relies on the graph Laplacian of a dataset, a single target image feature is insufficient to generate its embedding. Therefore, the client concatenates $\mathbf{r}$ with the public dataset $\mathcal{P}$. The client runs one iteration of PrivateMail where noise is added via the Gaussian mechanism before recomputing the Laplacian over the private embedding. This makes the next iterations that we run to be differentially private due to the post-processing invariance property as the iteration is now functionally independent of the raw features. We then run post-processing embeddings for a varying number of iterations depending on the dataset. Furthermore, since the client and server have access to different data, the embedding of $\mathbf{r} \cup \mathcal{P}$ on the client is not guaranteed to align with that of $\mathcal{S}$ on the server. We thus also concatenate $\mathcal{S}$ with $\mathcal{P}$ so the public data serves as a common ``anchor" for the embeddings, which is used to align the the embeddings of $\mathbf{r}$ and $\mathcal{S}$ via the Kabsch-Umeyama rigid-transformation algorithm \citep{umeyama1991least}. Once the server retrieves the $k$-nearest neighbors of the client's privatized embedding of $\mathbf{r}$ with respect to the server's non-private embedding of $\mathcal{S}$, the server gains additional information about $\mathbf{r}$ based on its neighbors. To obfuscate $\mathbf{r}$, we append a dataset $\mathcal{P}_\mathbf{r}$ of dummy queries to $\mathbf{r} \cup \mathcal{P}$ on the client-side. $\mathcal{P}_\mathbf{r}$ is generated by uniformly sampling images from the public dataset such that $\mathcal{P}_\mathbf{r}$ contains one image of every class besides the class of $\mathbf{r}$. The client's target image class is equally likely to be any of the possible classes in the dataset, so the server cannot directly infer the target class. The client is then able to filter out the retrieved images for the dummy targets. This process is visualized in Figure 1 and described in greater detail in Algorithm 1.

\begin{figure*}
        \centering \includegraphics[scale=0.55]{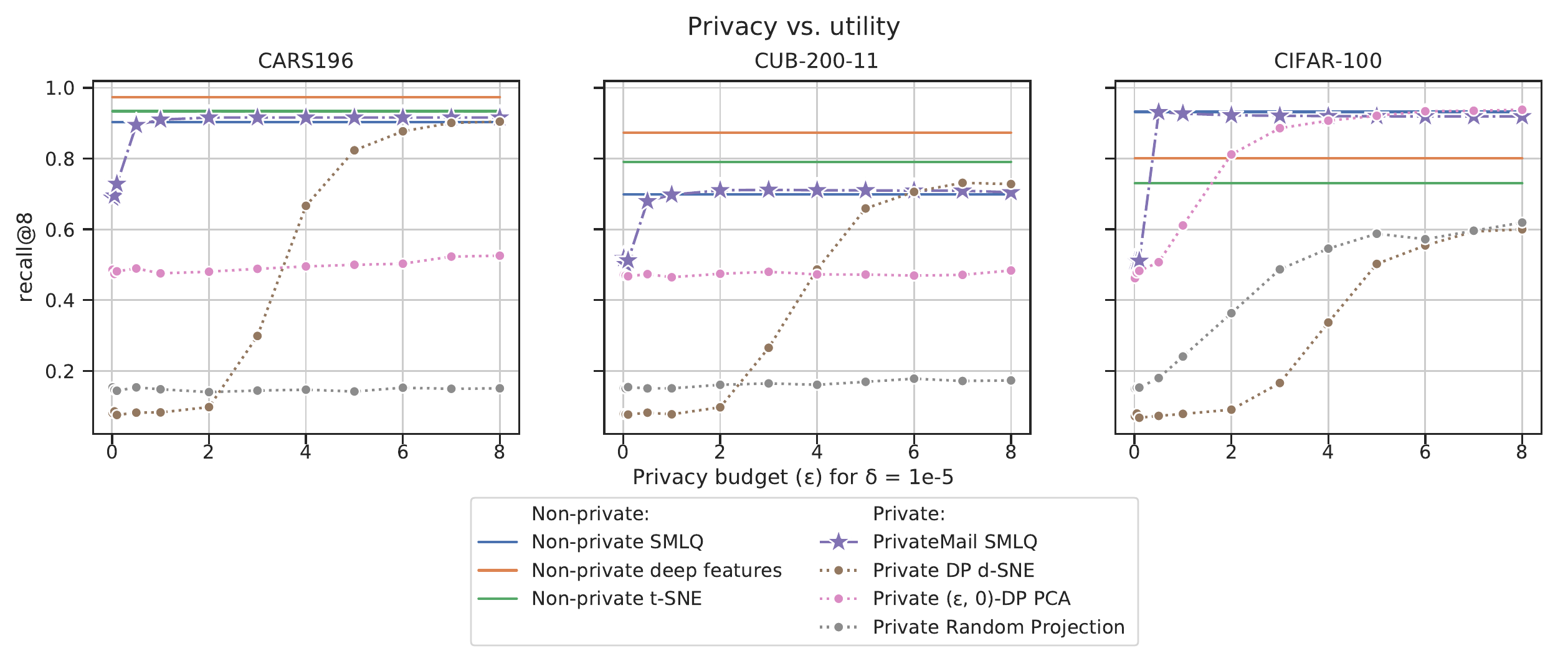}
        \caption{We compare the privacy-utility trade-off of PrivateMail with Recall@k experiments with $k=8$ for three datasets on 6 baselines that include private and non-private methods. The lower values of $\epsilon$ refer to higher levels of privacy. 
        }
        \label{fig:4}
    \end{figure*}
    
\begin{algorithm}[H]
\textbf{Input:}
Query $\mathbf{r}$, requested number of retrieved images $k$, number of post-processing iterations $T$.\\
\textbf{Output:} Server returns $k$ nearest matches w.r.t $\mathcal{S}$.\\
\textbf{Feature extraction:} Client extracts image-retrieval features $\mathbf{X}_\mathbf{r}$, $\mathbf{X}_\mathcal{P}$, $\mathbf{X}_{\mathcal{P}_\mathbf{r}}$ for $\mathbf{r}$, $\mathcal{P}$, $\mathcal{P}_\mathbf{r}$ from trained ML model. Server extracts features $\mathbf{X}_\mathcal{P}, \mathbf{X}_{\mathcal{S}}$ for $\mathcal{P}$,$\mathcal{S}$.\;\\
\textbf{Obfuscation:} Client concatenates $\mathbf{X}_\mathbf{r} \cup \mathbf{X}_{\mathcal{P}_\mathbf{r}}$ and labels. \;\\
\textbf{Anchoring with public data}: Client concatenates $\mathbf{X}_{\operatorname{client}} = \{\mathbf{X}_\mathbf{r} \cup \mathbf{X}_{\mathcal{P}_\mathbf{r}}\} \cup \mathbf{X}_\mathcal{P}$ and corresponding labels. Server concatenates $\mathbf{X}_{\operatorname{server}} = \mathbf{X}_\mathcal{S} \cup \mathbf{X}_\mathcal{P}$ and corresponding labels.\;\\
\textbf{Privatization}: Client runs \textbf{PrivateMail} mechanism on $\mathbf{X}_{\operatorname{client}}$ and $\mathbf{Y}_{\operatorname{client}}$ for 1 iter to obtain embedding $\mathbf{X}_{\operatorname{client}}'$.\;\\
\For{$t=1$ \KwTo $T$}{
 Client only runs step 2 of \textbf{PrivateMail} on $\mathbf{X}_{\operatorname{client}}'$ (using $L_{\mathbf{X}_{\operatorname{client}}'}$) to update the embeddings.\;\\
 Server only runs step 2 of \textbf{PrivateMail} on $\mathbf{X}_{\operatorname{server}}$ to obtain embedding $\mathbf{X}_{\operatorname{server}}'$.\;
}
\textbf{Align:} Non-private server embeddings and privatized client embeddings are aligned at server using Kabsch-Umeyama algorithm \cite{umeyama1991least}\;

\textbf{Retrieve:} Server retrieves $k$ nearest matches for each embedding of $\mathbf{r} \cup \mathcal{P}_\mathbf{r}$ in aligned dataset and serves to the client. \;

\textbf{Result parsing:} Client locates retrieved images for $\mathbf{r}$.

\caption{Differentially Private Image Retrieval}
\end{algorithm}

\section{Experiments}
\subsubsection{Datasets}
In this section we present experimental results on three important image retrieval benchmark datasets of i) Caltech-UCSD Birds-200-2011 (CUB-200-2011) \cite{WelinderEtal2010}, ii) Cars196 \cite{KrauseStarkDengFei-Fei_3DRR2013}, and iii) CIFAR-100 \cite{krizhevsky2009learning}.
\subsubsection{Methodology}
We use the state-of-the-art image retrieval method of \textit{`combination of multiple global descriptors'} (CGD) \citep{jun2019combination} with ResNet-50 \citep{He2015} backbone to generate features for the Cars196 and CUB-200-2011 datasets. CIFAR-100 features are extracted directly from ResNet-50 pre-trained on ImageNet \citep{deng2009imagenet}. We run Algorithm 1 on each dataset with the parameters outlined in appendix A. \subsubsection{Quantitative metrics}
We measure retrieval performance using the Recall@k metric as used in this popular non-private image retrieval paper \cite{jun2019combination}. As our proposed work is a differentially private algorithm, we study the \textit{utility-privacy trade-off} by looking at the recalls obtained at varying levels of $\epsilon$. Note that lower $\epsilon$ refers to higher privacy. 

\subsection{Baselines} We compare utility of our proposed PrivateMail mechanism against several important baselines as below.
    \\ \textbf{Non-private state of the art for image retrieval} We compare against the non-private method of CGD 
    that unfortunately does not preserve privacy, and see how close we get to its performance while also preserving privacy. Note that there exists a trade-off of privacy vs utility and the main goal is to preserve privacy, while attempting to maximize utility.\\\textbf{Differentially private unsupervised manifold embedding} A comparison with differentially private unsupervised manifold embedding method of DP-dSNE \cite{saha2020dsne,saha2021privacy} is done as this is one of the most recent manifold embedding methods with differential privacy. 
       \\ \textbf{Non-private supervised manifold embedding} We compare against non-private supervised manifold embedding to show how close our differentially private version fares in terms of achievable utility when the privacy is not at all preserved. 
        \\  \textbf{Non-private unsupervised manifold embedding} We compare against non-private unsupervised manifold embedding method of t-SNE \cite{van2008visualizing} to show the benefit of a supervised manifold embedding over an unsupervised embedding in terms of the utility.
         \\ \textbf{Differentially private classical projections} We compare against differentially private versions of more classical methods such as private PCA \cite{chaudhuri2013near} and private random projections \cite{kenthapadi2012privacy}. 

\subsection{Evaluation}
As shown in Figure \ref{fig:4}, PrivateMail SMLQ obtains a substantially better privacy-utility trade-off over all the considered private baselines on all the datasets. It also reaches closer to the methods that do not preserve privacy on CARS196. It even meets the non-private performance on CIFAR-100 at much higher levels of privacy (lower $\epsilon$'s). DP-dSNE reaches the performance of PrivateMail only at low levels of privacy on 2 out of the 3 datasets, while PrivateMail does substantially better at high-levels of privacy preservation. A similar phenomenon happens again with respect to private PCA on CIFAR-100.
\subsubsection{Effect of $k,\alpha, \sigma$}
In Figure \ref{fig:5}, we study the sensitivity of our method's performance with respect to various parameters such as choice of embedding dimension $k$, the weighting parameter $\alpha$ which acts as a regularizer for the embedding by weighting the graph Laplacians in the term $\mathbf{L_X} - \alpha \mathbf{L_Y}$ in our embedding update, and the $\sigma$ parameter used in defining the Gaussian kernels used to build $\mathbf{L_X,L_Y}$. As shown, tuning of $k,\alpha$ is stable while tuning of $\sigma$ requires a bit of a grid search. However, since we are in the supervised setting, standard methods for tuning could be used for practical purposes.
\subsubsection{Qualitative visualizations} 
\begin{figure}
\centering
    \includegraphics[width=0.69\linewidth]{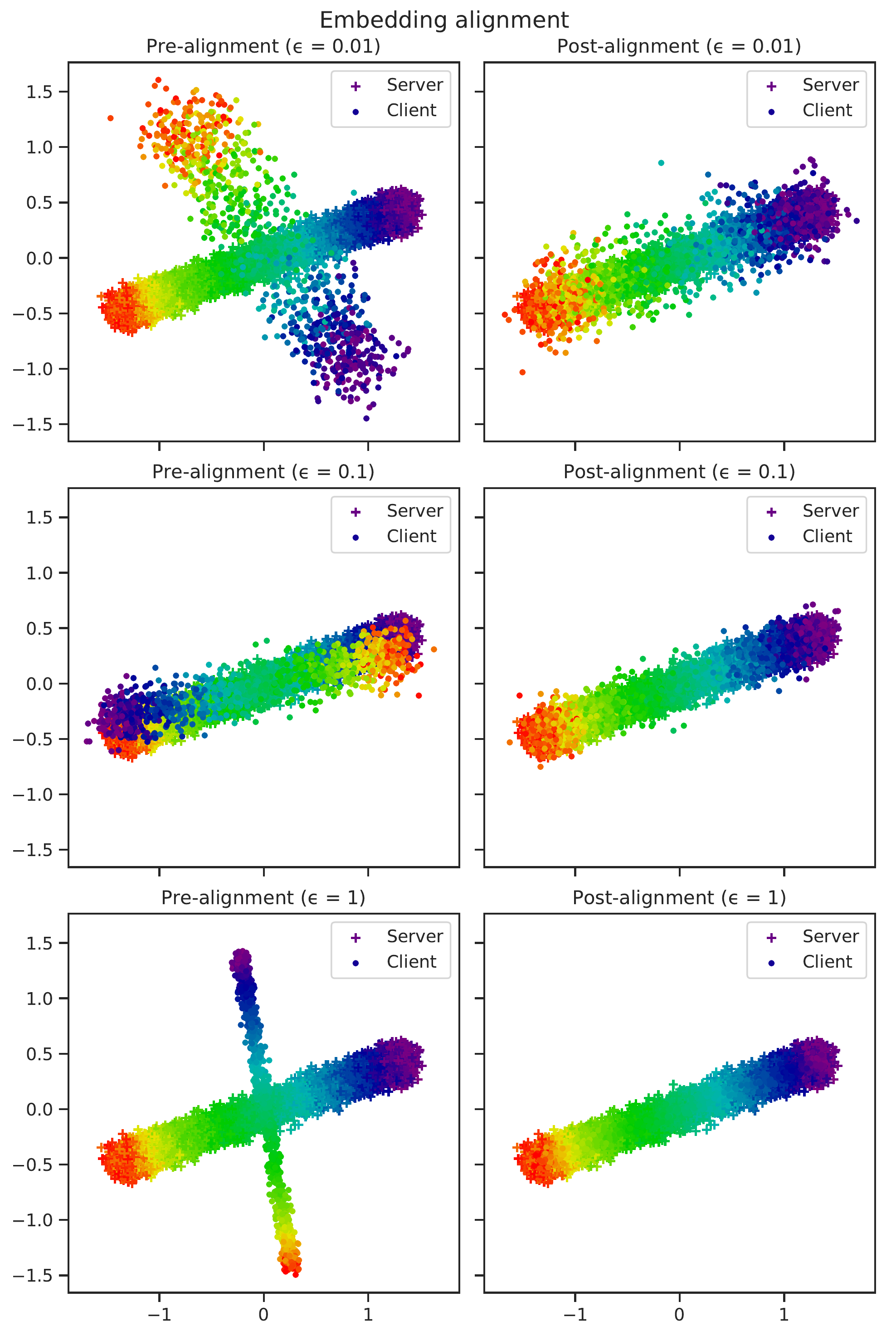}
    \caption{Embeddings for CARS196 data (with $\alpha=0.5$ and parameters in appendix A) at varying privacy levels $\epsilon$. We show that alignment improves as less noise is added. The privacy induced noise can be seen at various levels of $\epsilon$. 
   } 
   \label{fig:7}
\end{figure}

Example of  PrivateMail embeddings are given in Figure \ref{fig:7} for different values of privacy parameter $\epsilon$ pre- and post- server-client alignment. 


\section{Conclusion} We proposed a differentially private supervised manifold learning method and applied it to the private image retrieval problem. That said, there are a broad range of applications for manifold learning beyond that of image retrieval. Therefore, it would be interesting to investigate the potential benefits of doing these other tasks in a privacy preserving manner. We would like to extend the derived global sensitivity results to smooth sensitivities \cite{nissim2007smooth} in order to potentially further improve the privacy-utility trade-off.

\begin{quote}
\begin{small}
\bibliography{uai2021-template.bib}
\end{small}
\end{quote}

\onecolumn
\appendix

\section{Experiment parameters}\label{appendix:params}
Unless noted otherwise, we use the following parameters for the SMLQ experiments.
\begin{table}[!htbp]
\centering
\begin{tabular}{|l|l|l|l|}
\hline
\textbf{Parameter}                 & \textbf{CARS196} &
\textbf{CUB-200-2011} &
\textbf{CIFAR-100} 
\\ \hline
$\sigma$ & 6 &  5 & 6\\ \hline
$\alpha$ & 0.6 &  0.5 & 0.6\\  \hline
$k$ & 2 &  2 & 2\\          
\hline
$\sigma_q$ & $10^{-8}$ &  $10^{-8}$ & $10^{-8}$\\
\hline
$T$ & 5 &  5 & 5\\
\hline
$\epsilon$ & 0.1 &  0.1 & 0.1\\
\hline
$\delta$ & $10^{-5}$ & $10^{-5}$ & $10^{-5}$\\
\hline
\end{tabular}
\caption{Default experiment parameters}
\end{table}

\section{Global sensitivity derivation}\label{appendix:GS}

The proof of Theorem \ref{thm:GS} relies on deriving a constant bound $M$ such that $M \geq \|\mathbf{M_i}\|$ for all $\mathbf{X}$ and $\mathbf{\Tilde{X}}$, where $\mathbf{M_i}$ is the $i$th row of $\mathbf{Z}$ as defined in equation \eqref{eq:Z}.

\begin{lemma}
    For all $\mathbf{X}$ and $\mathbf{\Tilde{X}}$, $M = nM_{ij} + M_{ii} \geq \|\mathbf{M_i}\|$, where $M_{ii}^{\max}$ and $M_{ij}^{\max}$ are defined as
    \begin{gather}
        M_{ii}=\alpha^2\left[\left(\frac{n}{n e^{-\frac{2}{\sigma^2}} + e^{-\frac{1}{2\sigma^2}} - 1}\right)^2 + \left(\frac{n}{(n+1)e^{-\frac{2}{\sigma^2}} - 1}\right)^2 - \frac{2\left((n+1)e^{-\frac{c^2}{2\sigma^2}}-1 \right)^2}{n\left(n + e^{-\frac{1}{2\sigma^2}} - 1\right)}\right]\\
        M_{ij} = \frac{\alpha^2+1}{\left(ne^{-\frac{2}{\sigma^2}} + e^{-\frac{1}{2\sigma^2}}-1\right)^2}-\frac{2\alpha e^{-\frac{c^2+4}{2\sigma^2}} }{\left(n+e^{-\frac{1}{2\sigma^2}}-1\right)^2}
        + \frac{\alpha^2+1}{\left((n+1)e^{-\frac{2}{\sigma^2}}-1\right)^2}-\frac{2\alpha e^{-\frac{c^2+4}{2\sigma^2}}}{n^2} \nonumber\\
        -2\cdot\frac{\alpha^2 e^{-\frac{c^2}{\sigma^2}}+e^{-\frac{4}{\sigma^2}}}{n\left(n+e^{-\frac{1}{2\sigma^2}}-1\right)}+\frac{4\alpha}{\left(ne^{-\frac{2}{\sigma^2}} + e^{-\frac{1}{2\sigma^2}}-1\right)\left((n+1)e^{-\frac{2}{\sigma^2}}-1\right)}
    \end{gather}
\end{lemma}
\begin{proof}
Recall that we denote the $i$th row of $\mathbf{X}$ by $\mathbf{X_i}$. $\mathbf{L_{\Tilde{X}}}$, $\mathbf{L_Y}$, and $\mathbf{L_{\Tilde{Y}}}$ are defined similarly for $\mathbf{\Tilde{X}}$, $\mathbf{Y}$, and $\mathbf{\Tilde{Y}}$ respectively. Expanding the definition of $\mathbf{M_i}$,
\begin{align*}
    \|\mathbf{M_i}\| &= \sum_{j=1}^{n+1}\left[\Diag(\mathbf{L_X}){\dagger}\left[\alpha \mathbf{L_Y} - \mathbf{L_X} \right] - \Diag(\mathbf{L_{\Tilde{X}}}){\dagger}\left[\alpha \mathbf{L_{\Tilde{Y}}} - \mathbf{L_{\Tilde{X}}}\right]\right]_\mathbf{i,j}^2 \\
    &= \sum_{j=1}^{n+1}\left(\left[\Diag(\mathbf{L_X}){\dagger}\left[\alpha \mathbf{L_Y} - \mathbf{L_X} \right]\right]_\mathbf{i,j} - \left[\Diag(\mathbf{L_{\Tilde{X}}}){\dagger}\left[\alpha \mathbf{L_{\Tilde{Y}}} - \mathbf{L_{\Tilde{X}}}\right]\right]_\mathbf{i,j}\right)^2 \\
    &= \sum_{j=1}^{n+1}\left(
    \begin{aligned}
    &\left[\Diag(\mathbf{L_X}){\dagger}\left[\alpha \mathbf{L_Y} - \mathbf{L_X} \right]\right]_\mathbf{i,j}^2 + \left[\Diag(\mathbf{L_{\Tilde{X}}}){\dagger}\left[\alpha \mathbf{L_{\Tilde{Y}}} - \mathbf{L_{\Tilde{X}}}\right]\right]_\mathbf{i,j}^2 \\
    &- 2\left[\Diag(\mathbf{L_X}){\dagger}\left[\alpha \mathbf{L_Y} - \mathbf{L_X} \right]\right]_\mathbf{i,j}\left[\Diag(\mathbf{L_{\Tilde{X}}}){\dagger}\left[\alpha \mathbf{L_{\Tilde{Y}}} - \mathbf{L_{\Tilde{X}}}\right]\right]_\mathbf{i,j}
    \end{aligned}
    \right)
\end{align*}
Since the off-diagonal entries of $\Diag(\mathbf{L_X})$ and $\Diag(\mathbf{L_{\Tilde{X}}})$ are zero, we have
\begin{align*}
    \left[\Diag(\mathbf{L_X}){\dagger}\left[\alpha \mathbf{L_Y} - \mathbf{L_X} \right]\right]_\mathbf{i,j} &= \sum_{k=1}^n {\Diag(\mathbf{L_X}){\dagger}}_\mathbf{i,k}\left[\alpha \mathbf{L_Y} - \mathbf{L_X} \right]_\mathbf{k,j} = {\Diag(\mathbf{L_X}){\dagger}}_\mathbf{i,i}\left[\alpha \mathbf{L_Y} - \mathbf{L_X} \right]_\mathbf{i,j} = \frac{\alpha \mathbf{L_Y}_\mathbf{i,j} - \mathbf{L_X}_\mathbf{i,j} }{\mathbf{L_X}_\mathbf{i,i}}\\
    \left[\Diag(\mathbf{L_{\Tilde{X}}}){\dagger}\left[\alpha \mathbf{L_{\Tilde{Y}}} - \mathbf{L_{\Tilde{X}}} \right]\right]_\mathbf{i,j} &=  \frac{\alpha \mathbf{L_{\Tilde{Y}}}_\mathbf{i,j}  - \mathbf{L_{\Tilde{X}}}_\mathbf{i,j}}{\mathbf{L_{\Tilde{X}}}_\mathbf{i,i}}\
\end{align*}
Therefore, the norm of $\mathbf{M_i}$ is given by
\begin{equation}\label{eq:|M_i|}
    \|\mathbf{M_i}\| = \sum_{j=1}^{n+1} \left(\left(\frac{\alpha \mathbf{L_Y}_\mathbf{i,j}   - \mathbf{L_X}_\mathbf{i,j}  }{\mathbf{L_X}_\mathbf{i,i} }\right)^2 + \left( \frac{\alpha \mathbf{L_{\Tilde{Y}}}_\mathbf{i,j}   - \mathbf{L_{\Tilde{X}}}_\mathbf{i,j}  }{\mathbf{L_{\Tilde{X}}}_\mathbf{i,i}}\right)^2 - 2\frac{\left(\alpha \mathbf{L_Y}_\mathbf{i,j}   - \mathbf{L_X}_\mathbf{i,j}  \right)\left(\alpha \mathbf{L_{\Tilde{Y}}}_\mathbf{i,j}   - \mathbf{L_{\Tilde{X}}}_\mathbf{i,j}  \right)}{\mathbf{L_X}_\mathbf{i,i} \mathbf{L_{\Tilde{X}}}_\mathbf{i,i}}\right)
\end{equation}
We bound the above summation by bounding each summand,
\begin{equation}\label{eq:M_ij}
    M_{ij} = \left(\frac{\alpha \mathbf{L_Y}_\mathbf{i,j}   - \mathbf{L_X}_\mathbf{i,j}  }{\mathbf{L_X}_\mathbf{i,i} }\right)^2 + \left( \frac{\alpha \mathbf{L_{\Tilde{Y}}}_\mathbf{i,j}   - \mathbf{L_{\Tilde{X}}}_\mathbf{i,j}  }{\mathbf{L_{\Tilde{X}}}_\mathbf{i,i}}\right)^2 - 2\frac{\left(\alpha \mathbf{L_Y}_\mathbf{i,j}   - \mathbf{L_X}_\mathbf{i,j}  \right)\left(\alpha \mathbf{L_{\Tilde{Y}}}_\mathbf{i,j}   - \mathbf{L_{\Tilde{X}}}_\mathbf{i,j}  \right)}{\mathbf{L_X}_\mathbf{i,i} \mathbf{L_{\Tilde{X}}}_\mathbf{i,i}}
\end{equation}
Recall that the $(n+1)$th row of $\mathbf{X}$ and $\mathbf{Y}$ is $\mathbf{0}$. By the definition of the Laplacian in \eqref{LEqn1},
\begin{gather}
    \mathbf{{L_X}_{i,j}} = 
    \begin{cases}
        \sum_{k=1}^n \exp\left(-\frac{\|\mathbf{{X}_i}-\mathbf{{X}_k}\|^2}{2\sigma^2}\right) + \exp\left(-\frac{\| \mathbf{{X}_i}\|^2}{2\sigma^2}\right) - 1 & \text{if $i = j$}\\
        -\exp\left(-\frac{\|\mathbf{{X}_i}-\mathbf{{X}_j}\|^2}{2\sigma^2}\right) & \text{otherwise}
    \end{cases}\label{eq:L_X_ij}\\
    \mathbf{{L_Y}_{i,j}} = 
    \begin{cases}
        \sum_{k=1}^n \exp\left(-\frac{\|\mathbf{{Y}_i}-\mathbf{{Y}_k}\|^2}{2\sigma^2}\right) + \exp\left(-\frac{\| \mathbf{{Y}_i}\|^2}{2\sigma^2}\right) - 1 & \text{if $i = j$}\\
        -\exp\left(-\frac{\|\mathbf{{Y}_i}-\mathbf{{Y}_j}\|^2}{2\sigma^2}\right) & \text{otherwise}
    \end{cases}\label{eq:L_Y_ij}
\end{gather}
Let $\mathbf{v_X}$ and $\mathbf{v_Y}$ be the additional records in the $(n+1)$th rows of $\mathbf{L_{\Tilde{X}}}$ and $\mathbf{L_{\Tilde{Y}}}$ respectively. Then similarly to the above definitions of $\mathbf{{L_X}_{i,j}}$ and $\mathbf{{L_Y}_{i,j}}$, we have
\begin{gather}
    \mathbf{{L_{\Tilde{X}}}_{i,j}} = 
    \begin{cases}
        \sum_{k=1}^n \exp\left(-\frac{\|\mathbf{{X}_i}-\mathbf{{X}_k}\|^2}{2\sigma^2}\right) + \exp\left(-\frac{\| \mathbf{{{\Tilde{X}}}_i}-\mathbf{v_X}\|^2}{2\sigma^2}\right) - 1 & \text{if $i = j$}\\
        -\exp\left(-\frac{\|\mathbf{\Tilde{X}_i}-\mathbf{\Tilde{X}_j}\|^2}{2\sigma^2}\right) & \text{if $i \neq j$}
    \end{cases}\label{eq:L_Xtil_ij}\\
    \mathbf{{L_{\Tilde{Y}}}_{i,j}} = 
    \begin{cases}
        \sum_{k=1}^n \exp\left(-\frac{\|\mathbf{{Y}_i}-\mathbf{{Y}_k}\|^2}{2\sigma^2}\right) + \exp\left(-\frac{\| \mathbf{{{\Tilde{Y}}}_i}-\mathbf{v_Y}\|^2}{2\sigma^2}\right) - 1 & \text{if $i = j$}\\
        -\exp\left(-\frac{\|\mathbf{\Tilde{Y}_i}-\mathbf{\Tilde{Y}_j}\|^2}{2\sigma^2}\right) & \text{if $i \neq j$}
    \end{cases}\label{eq:L_Ytil_ij}
\end{gather}
We proceed to find upper and lower bounds for $M_{ij}$ by separately analyzing two cases: when $i=j$ and when $i\neq j$.
\begin{description}
    \item[Case 1: $i=j$.]
    By equation \eqref{eq:M_ij}, we have
    \begin{align}
        M_{ii} = \left(\frac{\alpha \mathbf{L_Y}_\mathbf{i,i}   - \mathbf{L_X}_\mathbf{i,i}  }{\mathbf{L_X}_\mathbf{i,i} }\right)^2 + \left( \frac{\alpha \mathbf{L_{\Tilde{Y}}}_\mathbf{i,i}   - \mathbf{L_{\Tilde{X}}}_\mathbf{i,i}  }{\mathbf{L_{\Tilde{X}}}_\mathbf{i,i}}\right)^2 - 2\frac{\left(\alpha \mathbf{L_Y}_\mathbf{i,i}   - \mathbf{L_X}_\mathbf{i,i}  \right)\left(\alpha \mathbf{L_{\Tilde{Y}}}_\mathbf{i,i}   - \mathbf{L_{\Tilde{X}}}_\mathbf{i,i}  \right)}{\mathbf{L_X}_\mathbf{i,i} \mathbf{L_{\Tilde{X}}}_\mathbf{i,i}}
    \end{align}
    This equation further simplifies to
    \begin{align}\label{eq:M_ii}
        M_{ii} &= \begin{aligned}[t]
        &\frac{\alpha^2\mathbf{L_Y}_\mathbf{i,i}^2-2\alpha\mathbf{L_X}_\mathbf{i,i} \mathbf{L_Y}_\mathbf{i,i}+\mathbf{L_X}_\mathbf{i,i}^2}{\mathbf{L_X}_\mathbf{i,i}^2}
        + \frac{\alpha^2{\mathbf{L_{\Tilde{Y}}}}_\mathbf{i,i}^2-2\alpha{\mathbf{L_{\Tilde{X}}}}_\mathbf{i,i} \mathbf{L_{\Tilde{Y}}}_\mathbf{i,i}+\mathbf{L_{\Tilde{X}}}_\mathbf{i,i}^2}{\mathbf{L_{\Tilde{X}}}_\mathbf{i,i}^2}\\
        &-2\frac{\alpha^2\mathbf{L_Y}_\mathbf{i,i}\mathbf{L_{\Tilde{Y}}}_\mathbf{i,i}-\alpha\mathbf{L_{\Tilde{X}}}_\mathbf{i,i} \mathbf{L_Y}_\mathbf{i,i}-\alpha\mathbf{L_X}_\mathbf{i,i} \mathbf{L_{\Tilde{Y}}}_\mathbf{i,i}+\mathbf{L_X}_\mathbf{i,i} \mathbf{L_{\Tilde{X}}}_\mathbf{i,i}}{\mathbf{L_X}_\mathbf{i,i} \mathbf{L_{\Tilde{X}}}_\mathbf{i,i}}
        \end{aligned}\nonumber\\
        &= \alpha^2 \left(\frac{\mathbf{L_Y}_\mathbf{i,i}^2}{\mathbf{L_X}_\mathbf{i,i}^2}+\frac{\mathbf{L_{\Tilde{Y}}}_\mathbf{i,i}^2}{\mathbf{L_{\Tilde{X}}}_\mathbf{i,i}^2} - \frac{2\mathbf{L_Y}_\mathbf{i,i}\mathbf{L_{\Tilde{Y}}}_\mathbf{i,i}}{\mathbf{L_X}_\mathbf{i,i}\mathbf{L_{\Tilde{X}}}_\mathbf{i,i}}\right)
    \end{align}
    Since each row in $\mathbf{X}$, $\mathbf{\Tilde{X}}$ is unit norm, by the law of cosines, we have
    \begin{align}\label{eq:cosine_law}
    \begin{split}
        \|\mathbf{{X}_i}-\mathbf{{X}_k}\|^2 &= \|\mathbf{{X}_i}\|^2 + \|\mathbf{{X}_k}\|^2 - 2\|\mathbf{{X}_i}\|\|\mathbf{{X}_k}\|\cos\theta_{\mathbf{{X}_i}, \mathbf{{X}_k}}= 2 - 2\cos\theta_{\mathbf{{X}_i}, \mathbf{{X}_k}}\\
        \|\mathbf{{{\Tilde{X}}}_i}-\mathbf{{{\Tilde{X}}}_k}\|^2 &= 2 - 2\cos\theta_{\mathbf{{\Tilde{X}}_i}, \mathbf{{\Tilde{X}}_k}}
    \end{split}
    \end{align}
    The cosine of the angle between two unit vectors falls between $-1$ and $1$. We use this property to bound $\mathbf{L_{X}}_\mathbf{i,i}$,
    \begin{align}\label{eq:diag_bounds_X}
    \begin{split}
        \sum_{k=1}^n \exp\left(-\frac{2 - 2(-1)}{2\sigma^2}\right) + \exp\left(-\frac{1}{2\sigma^2}\right) - 1 &\leq \mathbf{L_X}_\mathbf{i,i} \leq  \sum_{k=1}^n \exp\left(-\frac{2 - 2(1)}{2\sigma^2}\right) + \exp\left(-\frac{1}{2\sigma^2}\right) - 1 \\
        n e^{-\frac{2}{\sigma^2}} + e^{-\frac{1}{2\sigma^2}} - 1 &\leq \mathbf{L_X}_\mathbf{i,i} \leq n + e^{-\frac{1}{2\sigma^2}} - 1
    \end{split}
    \end{align}
    as well as $\mathbf{L_{\Tilde{X}}}_\mathbf{i,i}$,
    \begin{align}\label{eq:diag_bounds_Xtil}
    \begin{split}
        \sum_{k=1}^n \exp\left(-\frac{2 - 2(-1)}{2\sigma^2}\right) + \exp\left(-\frac{2 - 2(-1)}{2\sigma^2}\right) - 1 &\leq \mathbf{L_{\Tilde{X}}}_\mathbf{i,i} \leq  \sum_{k=1}^n \exp\left(-\frac{2 - 2(1)}{2\sigma^2}\right) + \exp\left(-\frac{2 - 2(1)}{2\sigma^2}\right) - 1  \\
        (n+1)e^{-\frac{2}{\sigma^2}} - 1 &\leq \mathbf{L_{\Tilde{X}}}_\mathbf{i,i} \leq n\\
    \end{split}
    \end{align}
    $\mathbf{Y}, \mathbf{\Tilde{Y}}$ are vectors of integer labels in $\{0, \ldots, c\}$, where $c+1$ is the number of unique classes in the dataset. We then have the constraints $0 <= \|\mathbf{{Y}_i}-\mathbf{{Y}_j}\|^2 <= c^2$, which generate the following bounds for $\mathbf{L_{Y}}_\mathbf{i,i}$,
        \begin{align}
        \begin{split}
            \sum_{k=1}^n \exp\left(-\frac{c^2}{2\sigma^2}\right) + \exp\left(-\frac{c^2}{2\sigma^2}\right) - 1 &\leq \mathbf{L_Y}_\mathbf{i,i} \leq  \sum_{k=1}^n \exp\left(-\frac{0}{2\sigma^2}\right) + \exp\left(-\frac{0}{2\sigma^2}\right) - 1 \\
            (n+1)e^{-\frac{c^2}{2\sigma^2}}-1 &\leq \mathbf{L_Y}_\mathbf{i,i} \leq  n
        \end{split}
        \end{align}
        and similarly for $\mathbf{L_{\Tilde{Y}}}_\mathbf{i,i}$,
        \begin{align}
        \begin{split}
            \sum_{k=1}^n \exp\left(-\frac{c^2}{2\sigma^2}\right) + \exp\left(-\frac{c^2}{2\sigma^2}\right) - 1 &\leq \mathbf{L_{\Tilde{Y}}}_\mathbf{i,i} \leq  \sum_{k=1}^n \exp\left(-\frac{0}{2\sigma^2}\right) + \exp\left(-\frac{0}{2\sigma^2}\right) - 1 \\
            (n+1)e^{-\frac{c^2}{2\sigma^2}}-1 &\leq \mathbf{L_{\Tilde{Y}}}_\mathbf{i,i} \leq  n 
        \end{split}
        \end{align}
        Combining these bounds with those in equations \eqref{eq:diag_bounds_X} and \eqref{eq:diag_bounds_Xtil}, we bound $M_{ii}$ from above by
        \begin{align}
            M_{ii} &\leq \alpha^2\left[\left(\frac{n}{n e^{-\frac{2}{\sigma^2}} + e^{-\frac{1}{2\sigma^2}} - 1}\right)^2 + \left(\frac{n}{(n+1)e^{-\frac{2}{\sigma^2}} - 1}\right)^2 - \frac{2\left((n+1)e^{-\frac{c^2}{2\sigma^2}}-1 \right)^2}{n\left(n + e^{-\frac{1}{2\sigma^2}} - 1\right)}\right] \nonumber \\
            &= M_{ii}
        \end{align}
    Now that we have derived an upper bounds for summands of the form $M_{ii}$ in \eqref{eq:M_ij}, we bound $M_{ij}$ where $i \neq j$.
    \item[Case 2: $i\neq j$.] Expanding \eqref{eq:M_ij}, we have
    \begin{align}\label{eq:M_ij_expanded}
        M_{ij} = \begin{aligned}[t]
        &= \frac{\alpha^2\mathbf{L_Y}_\mathbf{i,j}^2+\mathbf{L_X}_\mathbf{i,j}^2}{\mathbf{L_X}_\mathbf{i,i}^2}-\frac{2\alpha\mathbf{L_X}_\mathbf{i,j} \mathbf{L_Y}_\mathbf{i,j}}{\mathbf{L_X}_\mathbf{i,i}^2}
        + \frac{\alpha^2{\mathbf{L_{\Tilde{Y}}}}_\mathbf{i,j}^2+\mathbf{L_{\Tilde{X}}}_\mathbf{i,j}^2}{\mathbf{L_{\Tilde{X}}}_\mathbf{i,i}^2}-\frac{2\alpha{\mathbf{L_{\Tilde{X}}}}_\mathbf{i,j} \mathbf{L_{\Tilde{Y}}}_\mathbf{i,j}}{\mathbf{L_{\Tilde{X}}}_\mathbf{i,i}^2}\\
        &-2\cdot\frac{\alpha^2\mathbf{L_Y}_\mathbf{i,j}\mathbf{L_{\Tilde{Y}}}_\mathbf{i,j}+\mathbf{L_X}_\mathbf{i,j} \mathbf{L_{\Tilde{X}}}_\mathbf{i,j}}{\mathbf{L_X}_\mathbf{i,i} \mathbf{L_{\Tilde{X}}}_\mathbf{i,i}}+2\cdot\frac{\alpha\mathbf{L_{\Tilde{X}}}_\mathbf{i,j} \mathbf{L_Y}_\mathbf{i,j}+\alpha\mathbf{L_X}_\mathbf{i,j} \mathbf{L_{\Tilde{Y}}}_\mathbf{i,j}}{\mathbf{L_X}_\mathbf{i,i} \mathbf{L_{\Tilde{X}}}_\mathbf{i,i}}
        \end{aligned}
    \end{align}
    Similarly to the previous case, we use the law of cosines in \eqref{eq:cosine_law} to bound each term in \eqref{eq:M_ij}. Recall that we have already derived bounds for $\mathbf{L_X}_\mathbf{i,i}$ and $\mathbf{L_{\Tilde{X}}}_\mathbf{i,i}$. The bounds for $\mathbf{L_X}_\mathbf{i,j}$ are given by
    \begin{align}\label{eq:diag_bounds_y_a}
        \begin{split}
            -\exp\left(-\frac{2-2(1)}{2\sigma^2}\right) &\leq \mathbf{L_X}_\mathbf{i,j} \leq  - \exp\left(-\frac{2-2(-1)}{2\sigma^2}\right)\\
            -1 &\leq \mathbf{L_X}_\mathbf{i,j} \leq -e^{-\frac{2}{\sigma^2}}
        \end{split}
        \end{align}
        and for $\mathbf{L_{\Tilde{X}}}_\mathbf{i,j}$ by
        \begin{align}
        \begin{split}
            -\exp\left(-\frac{2-2(1)}{2\sigma^2}\right) &\leq \mathbf{L_{\Tilde{Y}}}_\mathbf{i,j} \leq   - \exp\left(-\frac{2-2(-1)}{2\sigma^2}\right) \\
            -1 &\leq \mathbf{L_{\Tilde{X}}}_\mathbf{i,j} \leq -e^{-\frac{2}{\sigma^2}}
        \end{split}
        \end{align}
    By the constraint $0 \leq \|\mathbf{Y_i}-\mathbf{Y_j}\|^2 \leq c^2$, 
        bounds for $\mathbf{L_{Y}}_\mathbf{i,j}$ are given by,
        \begin{align}
        \begin{split}
            -\exp\left(-\frac{0}{2\sigma^2}\right) &\leq \mathbf{L_Y}_\mathbf{i,j} \leq  - \exp\left(-\frac{c^2}{2\sigma^2}\right)\\
            -1 &\leq \mathbf{L_Y}_\mathbf{i,j} \leq -e^{-\frac{c}{2\sigma^2}}
        \end{split}
        \end{align}
        and for $\mathbf{L_{\Tilde{Y}}}_\mathbf{i,i}$,
        \begin{align}
        \begin{split}
            -\exp\left(-\frac{0}{2\sigma^2}\right) &\leq \mathbf{L_{\Tilde{Y}}}_\mathbf{i,j} \leq   - \exp\left(-\frac{c^2}{2\sigma^2}\right) \\
            -1 &\leq \mathbf{L_{\Tilde{Y}}}_\mathbf{i,j} \leq -e^{-\frac{c}{2\sigma^2}}
        \end{split}
        \end{align}
        Substituting these bounds into \eqref{eq:M_ij_expanded}, we bound $M_{ij}$ from above by
        \begin{align}
        M_{ij} &\leq \begin{aligned}[t] &\frac{\alpha^2(-1)^2+(-1)^2}{(ne^{-\frac{2}{\sigma^2}} + e^{-\frac{1}{2\sigma^2}}-1)^2}-\frac{2\alpha(-e^{-\frac{2}{\sigma^2}}) (-e^{-\frac{c^2}{2\sigma^2}})}{(n+e^{-\frac{1}{2\sigma^2}}-1)^2}
        + \frac{\alpha^2(-1)^2+(-1)^2}{((n+1)e^{-\frac{2}{\sigma^2}}-1)^2}-\frac{2\alpha(-e^{-\frac{2}{\sigma^2}}) (-e^{-\frac{c^2}{2\sigma^2}})}{n^2}\\
        &-2\cdot\frac{\alpha^2(-e^{-\frac{c^2}{2\sigma^2}})^2+(-e^{-\frac{2}{\sigma^2}})^2}{n(n+e^{-\frac{1}{2\sigma^2}}-1)}+2\cdot\frac{\alpha(-1)^2+\alpha(-1)^2}{(ne^{-\frac{2}{\sigma^2}} + e^{-\frac{1}{2\sigma^2}}-1)((n+1)e^{-\frac{2}{\sigma^2}}-1)}
        \end{aligned}\nonumber\\
        &= \begin{aligned}[t] &\frac{\alpha^2+1}{(ne^{-\frac{2}{\sigma^2}} + e^{-\frac{1}{2\sigma^2}}-1)^2}-\frac{2\alpha e^{-\frac{c^2+4}{2\sigma^2}} }{(n+e^{-\frac{1}{2\sigma^2}}-1)^2}
        + \frac{\alpha^2+1}{((n+1)e^{-\frac{2}{\sigma^2}}-1)^2}-\frac{2\alpha e^{-\frac{c^2+4}{2\sigma^2}}}{n^2}\\
        &-2\cdot\frac{\alpha^2 e^{-\frac{c^2}{\sigma^2}}+e^{-\frac{4}{\sigma^2}}}{n(n+e^{-\frac{1}{2\sigma^2}}-1)}+\frac{4\alpha}{(ne^{-\frac{2}{\sigma^2}} + e^{-\frac{1}{2\sigma^2}}-1)((n+1)e^{-\frac{2}{\sigma^2}}-1)}
        \end{aligned} \nonumber\\
        &= M_{ij}^{\max}
        \end{align}
\end{description}
Therefore, an upper bound for $\|\mathbf{M_i}\|$ is given by
\begin{align}\label{eq:z^max}
    \|\mathbf{M_i}\| &\leq \sum_{j\neq i} M_{ij} + M_{ii} \nonumber\\
    &= nM_{ij} + M_{ii} \nonumber\\
    &= M 
\end{align}
\end{proof}
Applying this bound, which holds for all $\mathbf{X}$ and $\mathbf{\Tilde{X}}$, to equation \eqref{eq:GS_inter} in the proof of Theorem 2, we obtain a bound on the sensitivity of the SMLQ.

\section{Optimization} \label{AppdxA}
\textbf{Solution without matrix inverses or a step size parameter}
In this section we formulate an efficient monotonically convergent solution for the proposed supervised embedding loss where the update does not require a matrix inverse or a step size parameter. In empirical results we saw that even few iterations of our solution was good enough to give a great embedding. For brevity, we refer to the embedding $\mathbf{f(X)}$ by $\mathbf{Z}$ in this appendix.

\subsubsection{Concave-convex procedure: Special case of majorization minimization}
A function $g(\mathbf{Z}_{t+1}, \mathbf{X}_t)$ is said to majorize the function $v(\mathbf{Z})$ at
$\mathbf{Z}_{t}$ provided
$v(\mathbf{Z}_{t}) = g(\mathbf{Z}_{t},\mathbf{Z}_{t})$ and
$v(\mathbf{X}_{t}) \leq g(\mathbf{X}_{t},\mathbf{Z}_{t+1})$ always holds true. The MM iteration guarantees monotonic convergence \citep{hunter2004tutorial,wu2010mm,lange2016mm,zhou2019mm} because of this sandwich inequality that directly arises due to the above definition of majorization functions.$$v(\mathbf{Z}_{t+1}) \leq  g(\mathbf{Z}_{t+1},\mathbf{X}_t) \leq g(\mathbf{X}_t,\mathbf{X}_t) = v(\mathbf{X}_t)$$
The concave-convex procedure to solve the difference of convex (DC) optimization problems is a special case of MM algorithms as follows. 
For objective functions $v(\mathbf{X})$ which can be written as a difference of convex functions as $v_{vex}(\mathbf{Z}) + v_{cave}(\mathbf{Z})$ we have the following majorization function that satsifies the two properties described in the beginning of this subsection.

\begin{equation}
    v(\mathbf{Z})\leq v_{vex}(\mathbf{Z}) + v_{cave}(\mathbf{X})+(\mathbf{Z}-\mathbf{X})^T\nabla v_{cave}(\mathbf{X})=g(\mathbf{Z},\mathbf{X})
\end{equation} where $g(\mathbf{Z},\mathbf{Z}) = v(\mathbf{Z})$ and $g(\mathbf{Z},\mathbf{X}) \geq v(\mathbf{Z})$ when $\mathbf{Z}\neq \mathbf{X}$.

Therefore the majorization minimization iterations are \begin{enumerate}
    \item Solve   $\frac{\partial g(\mathbf{Z}_{t+1},\mathbf{X}_t)}{\partial \mathbf{Z}_t+1} = 0$ for $\mathbf{X}_t$
    \item Set $\mathbf{X}_t = \mathbf{Z}_t$ and continue till convergence.
\end{enumerate}

This gives the iteration known as the concave-convex procedure.
\begin{equation}
    \nabla v_{vex}(\mathbf{Z}_{t+1}) = -\nabla v_{cave}(\mathbf{Z}_t )
\end{equation}

\subsubsection{Iterative Update with Monotonic convergence for SMLQ}
\begin{proof}
We denote by $\Diag(\mathbf{L_X})$, a diagonal matrix whose diagonal is the diagonal of $\mathbf{L_X}$. Now, we can
build a majorization function over $\Tr{(\mathbf{X^TL_XX})}$, based on the fact that $2\Diag(\mathbf{L_X}) - \mathbf{L_X}$
is diagonally dominant. This leads to the following inequality for any matrix $\mathbf{M}$ with real entries and of the same dimension as $\mathbf{X}$.
$$\Tr(\mathbf{(X-M)^T[2\Diag(L_X) - L_X](X-M)})\geq 0$$ as also used in \citep{vepakomma2018supervised}.
We now get the following majorization inequality over $\Tr{(\mathbf{X^TL_XX})}$, by separating it from the above inequality.

\begin{align*} 
\mathbf{\Tr{(X^TL_XX)}+ b(Y)} \leq \mathbf{\Tr{[2X^T\Diag(L_X)X]}}&-\\2\Tr{\mathbf{[X^T(2\Diag(L_X)-L_X)M]}} = \lambda(\mathbf{X,M}) 
\end{align*}

which is quadratic in $\mathbf{X}$ where, $\mathbf{b(M)=\Tr{(M^TL_XM)} -\Tr{(M^T2\Diag(L_X)M)}}$. Let $h(\mathbf{X,M})=\lambda(\mathbf{X,M})-\alpha\Tr{(\mathbf{X^TL_YX})}$

This leads to the following bound over our loss function with $const(\mathbf{M})$ being a function that only depends on $\mathbf{M}$: 
\begin{align*}
\mathbf{G(X)}+const(\mathbf{M}) &\leq h(\mathbf{X,M}) \text{ } \forall \mathbf{X} \neq \mathbf{M}\\ &=h(\mathbf{X,X}) \text{, when } \mathbf{X}=\mathbf{M}
\end{align*} that satisfies the supporting point requirement, and hence $h(\cdot)$ touches the objective function at the current iterate and forms a majorization function. Now the following majorization minimization iteration holds true for an iteration $t$:

$$\mathbf{X}_{t+1} = \underset{\mathbf{X}}{\mathrm{argmin}} \text{ } h(\mathbf{X,M_t}) \text{ and } \mathbf{M_{t+1} = X_t}$$

It is important to note that these inequalities occur amongst the presence of additive terms, $\mathbf{const(M)}$ that are independent of $\mathbf{X}$ unlike a typical majorization-minimization framework and hence, it is a relaxation. The majorization function $\mathbf{h(X, M_t)}$ can be expressed as a sum of a convex function $e_{vex}(\mathbf{X}) = \lambda(\mathbf{X, M_t})$ and a concave function $e_{cave}(\mathbf{X}) = -\alpha \Tr{(\mathbf{X^TL_YX})}$. By the concave-convex formulation, we get the iterative solution by solving for $\nabla e_{vex}(\mathbf{X}_t) = -\nabla e_{cave}(\mathbf{X}_{t-1})$ which gives us
$$ \mathbf{X_t} = \frac{\alpha}{2}\Diag(\mathbf{L_X})^{\dagger}\mathbf{L_Y X_{t-1} + M_t} -\frac{1}{2} \Diag(\mathbf{L_X})^{\dagger}\mathbf{L_XM_t} $$ and on applying the majorization update over $\mathbf{M}_t$, we finally get the supervised manifold learning update that does not require a matrix inversion or a step-size parameter while guaranteeing monotonic convergence.
\end{proof}If the concave Hessian has small curvature
compared to the convex Hessian in the neighborhood
of an optima, then CCCP will generally have a superlinear convergence like quasi-Newton methods. This and other characterizations for convergence of CCCP, under various settings has been studied in great detail in \cite{salakhutdinov2012convergence}. 
\section{Separation-regularity trade-off in  supervised manifold learning \cite{vural2017study}}
\begin{theorem}
     Let $X=\left\{x_{i}\right\}_{i=1}^{N} \subset \mathbb{R}^{n}$ be a set of training samples such that each $x_{i}$ is drawn i.i.d. from one of the probability measures $\left\{\nu_{m}\right\}_{m=1}^{M}$, with $\nu_{m}$ denoting the probability measure of the $m$-th class. Let $Z=\left\{y_{i}\right\}_{i=1}^{N}$ be an embedding of $X$ in $\mathbb{R}^{d}$ such that there exist a constant $\gamma>0$ and a constant $A_{\delta}$ depending on $\delta>0$ satisfying
$$
\begin{aligned}
&\left\|z_{i}-z_{j}\right\|<A_{\delta}, \text { if }\left\|x_{i}-x_{j}\right\| \leqslant 2 \delta \text { and } C\left(x_{i}\right)=C\left(x_{j}\right) \\
&\left\|z_{i}-z_{j}\right\|>\gamma, \text { if } C\left(x_{i}\right) \neq C\left(x_{j}\right)
\end{aligned}
$$
For given $\epsilon>0$ and $\delta>0$, let $f: \mathbb{R}^{n} \rightarrow \mathbb{R}^{d}$ be a Lipschitz continuous interpolation function with constant $L$, which maps each $x_{i}$ to $f\left(x_{i}\right)=y_{i}$, such that
$$
L \delta+\sqrt{d} \epsilon+A_{\delta} \leqslant \frac{\gamma}{2}
$$
Consider a test sample $x$ randomly drawn according to the probability measure $\nu_{m}$ of class $m .$ For any $Q>0$, if $X$ contains at least $N_{m}$ training samples from the $m$-th class drawn i.i.d. from $\nu_{m}$ such that
$$
N_{m}>\frac{Q}{\eta_{m, \delta}}
$$
then the probability of correctly classifying $x$ with 1-NN classification in $\mathbb{R}^{d}$ is lower bounded as
$$
\begin{gathered}
P(\hat{C}(x)=m) \geq  1-\exp \left(-\frac{2\left(N_{m} \eta_{m, \delta}-Q\right)^{2}}{N_{m}}\right) 
-2 d \exp \left(-\frac{Q \epsilon^{2}}{2 L^{2} \delta^{2}}\right)
\end{gathered}
$$
\end{theorem}

\end{document}